\documentclass[10pt,twocolumn,letterpaper]{article}

\usepackage[pagenumbers]{cvpr} 

\usepackage{graphicx}
\usepackage{amsmath}
\usepackage{amssymb}
\usepackage{booktabs}
\usepackage{multirow}
\usepackage{adjustbox}
\usepackage{subcaption}
\usepackage[super]{nth}
\usepackage{pifont}
\usepackage{enumitem}
\newcommand{\floor}[1]{\lfloor #1 \rfloor}
\usepackage[dvipsnames]{xcolor}
\usepackage[pagebackref,breaklinks,colorlinks,citecolor=green]{hyperref}

\usepackage[capitalize]{cleveref}
\crefname{section}{Sec.}{Secs.}
\Crefname{section}{Section}{Sections}
\Crefname{table}{Table}{Tables}
\crefname{table}{Tab.}{Tabs.}

\def\cvprPaperID{983} 
\def\confName{CVPR}
\def\confYear{2023}

\begin{document}
	
	\title{MSF: Motion-guided Sequential Fusion for Efficient 3D \\ Object Detection from Point Cloud Sequences}
	
	\author{
		Chenhang He\thanks{Equal contribution.} \,\,\,\, Ruihuang Li$^*$ \,\, Yabin Zhang \,\, Shuai Li \,\, Lei Zhang\thanks{Corresponding author.} \\
		The Hong Kong Polytechnic University\\
		{\tt\small \{csche, csrhli, csybzhang, cssli, cslzhang\}@comp.polyu.edu.hk}}
	
	\maketitle
	
	\begin{abstract}
		Point cloud sequences are commonly used to accurately detect 3D objects in applications such as autonomous driving. Current top-performing multi-frame detectors mostly follow a Detect-and-Fuse framework, which extracts features from each frame of the sequence and fuses them to detect the objects in the current frame. However, this inevitably leads to redundant computation since adjacent frames are highly correlated. In this paper, we propose an efficient Motion-guided Sequential Fusion (MSF) method, which exploits the continuity of object motion to mine useful sequential contexts for object detection in the current frame. We first generate 3D proposals on the current frame and propagate them to preceding frames based on the estimated velocities. The points-of-interest are then pooled from the sequence and encoded as proposal features. A novel Bidirectional Feature Aggregation (BiFA) module is further proposed to facilitate the interactions of proposal features across frames. Besides, we optimize the point cloud pooling by a voxel-based sampling technique so that millions of points can be processed in several milliseconds. 
		The proposed MSF method achieves not only better efficiency than other multi-frame detectors but also leading accuracy, with 83.12\% and 78.30\% mAP on the LEVEL1 and LEVEL2 test sets of Waymo Open Dataset, respectively. Codes can be found at \url{https://github.com/skyhehe123/MSF}.
	\end{abstract}

	\section{Introduction}
	\label{sec:intro}
	
	\begin{figure}[t]
		
		\centering
		\includegraphics[width=0.99\columnwidth, height=5.5cm]{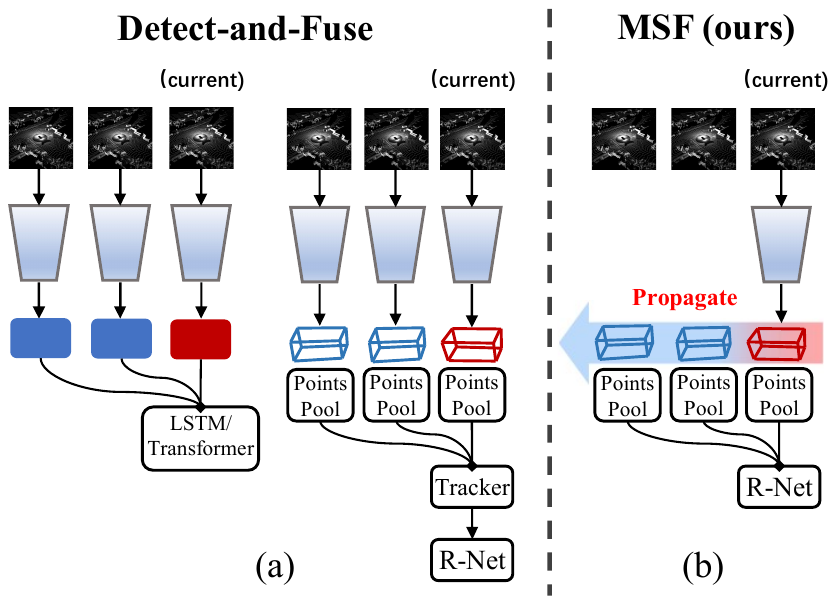}
		
		\vspace{-3mm}
		
		\caption{ (a) The ``Detect-and-Fuse'' framework extracts features from each frame of the sequence and then fuses them, while (b) our proposed ``Motion-guided Sequential Fusion" (MSF) method generates proposals on the current frame and propagates them to preceding frames to explore useful contexts in the sequence. 
		}
		\label{fig:motivation}
		\vspace{-5mm}
	\end{figure}
	
	3D object detection \cite{voxelnet,pointpillars,second,pvrcnn,mv3d,3dssd,sa-ssd,mppnet,voxset,votr,vote3d} is one of the key technologies in autonomous driving, which helps the vehicle to better understand the surrounding environment and make critical decisions in the downstream tasks. As an indispensable sensing device in autonomous driving systems, LiDAR collects 3D measurements of the scene in the form of point clouds. However, LiDAR can only produce partial view of the scene at a time, and the sparse and incomplete representation of point clouds brings considerable challenges to the 3D object detection task. In practice, the LiDAR sensor will continuously sense the environment and produce a sequence of point cloud frames over time. The multi-frame data can provide a denser representation of the scene as the vehicle moves. Therefore, how to fuse these multi-frame point cloud data for more accurate object detection is worth deep investigation. 
	
	Recent works mainly focus on deep feature fusion with multi-frame point clouds, for example, aggregating dense birds-eye-view features via Transformer models \cite{3dman, centerformer}, passing the voxel features to LSTM \cite{lstm3d} or GRU \cite{graph3d} modules for temporal modeling.  Some top-performing detectors \cite{offboard, mppnet} focus on fusing proposal features, where a tracker is employed to associate the 3D proposals across frames, and a region-based network is applied to refine the current proposals by incorporating contextual features from the proposal trajectories. These approaches generally follow a ``Detect-and-Fuse" framework, as shown in Fig.~\textcolor{red}{\ref{fig:motivation}(a)}, where the model requires to process each frame of the sequence, and the predictions on the current frame rely on the results of preceding frames. Since online detection is a causal system, such a detection framework might cause significant delay if the network is still processing a preceding frame when the current frame is loaded.
	
	In this paper, we propose an efficient Motion-guided Sequential Fusion (MSF) method, as shown in Fig.~\textcolor{red}{\ref{fig:motivation}(b)}, which leverages the continuity of object motion to extract useful contexts from point cloud sequences and improve the detection of current frame. Specifically, considering that the motions of objects are relatively smooth in a short sequence, we propagate the proposals generated on current frame to preceding frames based on the velocities of objects, and sample reliable points-of-interest from the sequence. In this way, we bypass extracting features on each frame of the sequence, which reduces the redundant computation and reliance on the results of preceding frames. The sampled points are then transformed to proposal features via two encoding schemes and passed to a region-based network for further refinement. Specifically, a self-attention module is employed to enhance the interaction of point features within proposals, while a novel Bidirectional Feature Aggregation (BiFA) module is proposed to enforce the information exchange between proposals across frames. The refined proposal features consequently capture both spatial details and long-term dependencies over the sequence, leading to more accurate bounding-box prediction.

	It is found that the existing point cloud pooling methods \cite{pointrcnn, std, ct3d, mppnet} are inefficient, taking more than 40 milliseconds when processing millions of points from sequential point clouds. We find that the major bottleneck lies in the heavy computation of pair-wise distances between $n$ points and $m$ proposals, which costs $\mathcal O(nm)$ complexity. To further improve the efficiency, we optimize the point cloud pooling with a voxel sampling technique. The improved pooling operation is of linear complexity and can process millions of points in several milliseconds, more than eight times faster than the original method. 
	
    Overall, our contributions can be summarized as follows.
 	\vspace{-4mm}
    \begin{itemize}
		\item[$\bullet$] An efficient Motion-guided Sequential Fusion (MSF) method is proposed to fuse multi-frame point clouds at region level by propagating the proposals of current frame to preceding frames based on the object motions.
		\vspace{-5mm}	
		\item[$\bullet$] A novel Bidirectional Feature Aggregation (BiFA) module is introduced to facilitate the interactions of proposal features across frames.
		
		\item[$\bullet$] The point cloud pooling method is optimized with a voxel-based sampling technique, significantly reducing the runtime on large-scale point cloud sequence.
    \end{itemize}
   
    The proposed MSF method is validated on the challenging Waymo Open Dataset, and it achieves leading accuracy on the LEVEL1 and LEVEL2 test sets with fast speed.
	
	\begin{figure*}[ht]
		\includegraphics[width=1\textwidth]{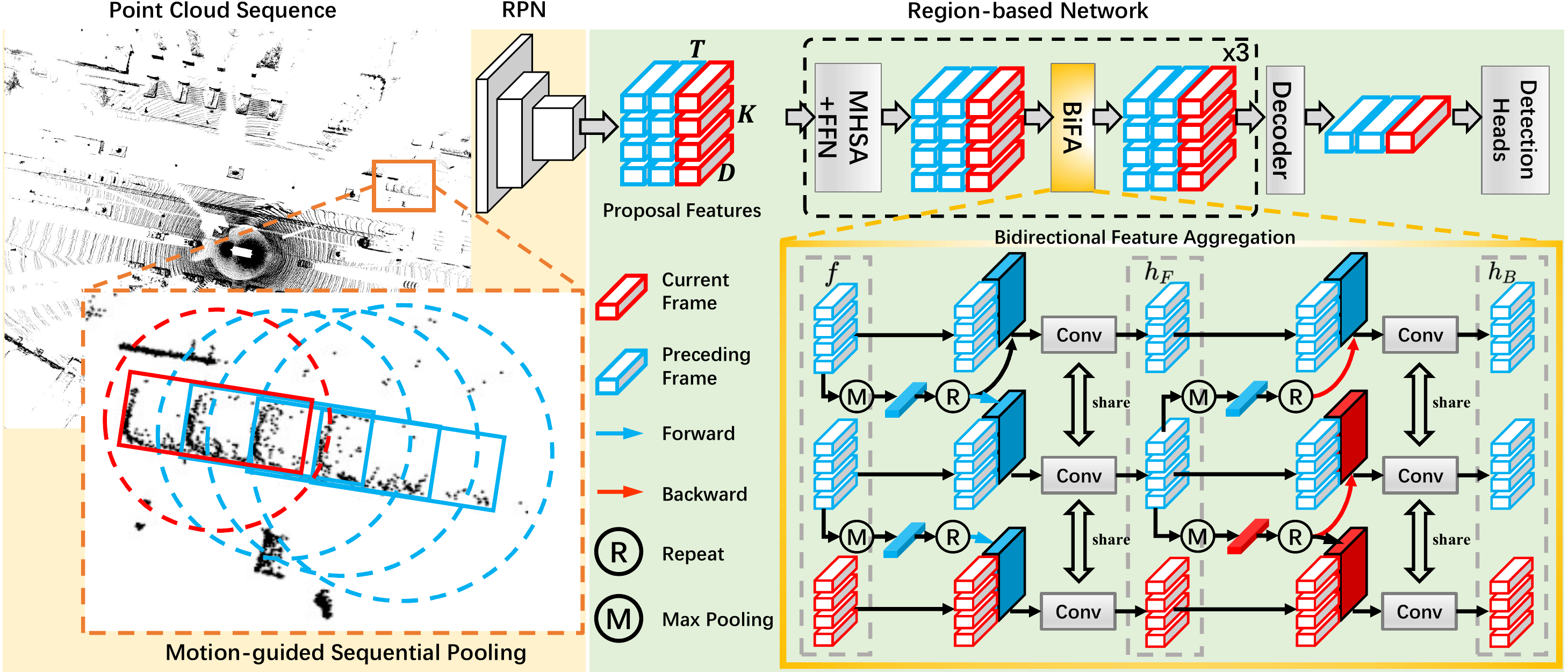}
		\caption{The overall architecture of our proposed Motion-guided Sequential Fusion (MSF) approach. By taking a point cloud sequence as input, MSF employs a region proposal network to generate proposals on the current frame and sample points-of-interest from the sequence by using motion-guided sequential pooling. The sampled points are encoded as high-dimensional proposal features and passed to a region-based network, where three learning blocks are consequently applied to refine the proposal features. A Bidirectional Feature Aggregation (BiFA) module is introduced in the region-based network to facilitate the interactions of proposal features across frames. The red and blue cubes represent single-point features from the current frame and preceding frame, respectively.}
		
		\label{fig:model}
	\end{figure*}
	
	\section{Related Work}
	\textbf{Single-frame 3D object detection.}
	Recent research on single-frame 3D object detection is mainly focused on representation learning on point clouds. Voxel-based detectors \cite{voxelnet, second, centerpoint} rasterize the point cloud into volumetric representation, followed by 3D CNN to extract dense features. Some works convert point clouds into 2D birds-eye-view \cite{pointpillars} or range view \cite{rangedet, rangeioudet} representations, and process them with more efficient 2D CNN. Following PointNet++ \cite{pointnet++}, point-based methods \cite{3dssd, std, pointrcnn, iassd, votenet} directly process point clouds in continuous space, and extract highly-semantic features through a series of downsampling and set abstraction layers. Voxel-point approaches \cite{sa-ssd, pvrcnn, pvcnn} employ a hybrid representation, where the flexible conversion between voxel-based and point-based representations are explored, leading to better balance between efficiency and performance. Our method employs a high-quality voxel-based detector CenterPoint~\cite{centerpoint} as the proposal generation network to predict 3D proposals of current frame and their motions. We then employ an efficient region-based network to further refine these proposals by mining sequential points from point cloud sequence.
	
	\textbf{3D object detection from point cloud sequence.}
	Multi-frame point clouds provide richer 3D information of the environment. While some single-frame detectors \cite{centerpoint, sst} can be adapted to point cloud sequence by simply concatenating multi-frame point cloud as the input, the improvements are typically marginal and the performance can be even worse when encountering moving objects. Fast-and-furious \cite{faf} explores an intermediate fusion to align multi-frame point cloud by concatenating the hidden feature maps of the backbone network. However, it still suffers from the misalignment brought by the fast-moving objects in long sequence. Recent approaches \cite{lstm3d, graph3d} demonstrate that an in-depth fusion can be achieved with recurrent networks. Unfortunately, the use of a single memory to store and update features across frames builds a potential bottleneck. To resolve such limitations, 3D-MAN \cite{3dman} first attempts to employ the attention mechanism to align different views of 3D objects and then exploits a memory bank to store and aggregate multi-frame features for long sequence. Recently, Offboard3D \cite{offboard} and MPPNet\cite{mppnet} improve much the detection performance, where they associate the detected boxes from each frame of the sequence as proposal trajectories, and extract high-quality proposal features by sampling sequential point cloud on the trajectories.
	
	Our MSF method also samples points from the sequence, but it differs from those methods with proposal trajectories \cite{mppnet, offboard} in that we only generate proposals on the current frame and propagate them to explore features in preceding frames. This makes our method much more efficient and favorable to online detection systems.

	\begin{table}[t]
		\caption{Recall rates of foreground points by using per-frame detection based proposal trajectory method \cite{mppnet} and our motion guided proposal generation method. We employ the CenterPoint \cite{centerpoint} as proposal generator and evaluate on Waymo validation split.}
		\centering
		\begin{adjustbox}{width=0.9\columnwidth}
			\begin{tabular}{c|ccc}
				\hline
				
				{} &  \textit{4-frame}  & \textit{8-frame}  & \textit{16-frame}  \\ \hline\hline
				Trajectory \cite{mppnet} &93.2\%& 92.8\%& 90.5\% \\
				Ours ($\gamma=1.0$) &92.3\% &87.5\%  &78.3\% \\
				Ours ($\gamma=1.1$) &93.5\% &91.7\%  &87.3\% \\
				\hline 
				
			\end{tabular}
		\end{adjustbox}
		\vspace{-5mm}
		\label{tbl:pp}	
	\end{table}

	\section{Motion-guided Sequential Fusion}
	This section presents our Motion-guided Sequential Fusion (MSF) approach for efficient 3D object detection on point cloud sequences. The overall architecture of MSF is illustrated in Fig.~\ref{fig:model}.  In Sec.~\textcolor{red}{3.1}, we describe the details of motion-guided sequential pooling, which effectively mines reliable sequential points-of-interest based on the proposals of current frame. In Sec. \textcolor{red}{3.2}, we present the region-based network, including the formulation of proposal features and a novel bidirectional feature aggregation module. In Sec. \textcolor{red}{3.3}, we demonstrate a voxel-based sampling technique to accelerate the current point cloud pooling method.

	\subsection{Motion-guided Sequential Pooling}
	Current multi-frame detection methods~\cite{offboard, mppnet} mostly explore proposal trajectories to generate high-quality point cloud representations. However, such a scheme relies on frame-by-frame proposal generation, which is not suitable for online detection systems. We observe that in a point cloud sequence, although objects move at different speeds, their motions are relatively smooth. That is to say, we can estimate their motion displacements and roughly localize their positions in preceding frames.
	To this end, given a point cloud sequence $\{I^t\}_{t=1}^{T}$, we propose to propagate the proposals generated on the current frame $I^T$ to preceding frames $\{I^t\}_{t=1}^{T-1}$ based on their estimated velocities. Since moving objects may slightly deviate from the estimated positions in the preceding frames, we sample the points-of-interest in a cylindrical region of each proposal and gradually increase the diameter of the region  by a factor $\gamma$ as the proposal propagates.
	Let's denote a proposal of current frame as $(p_x, p_y, p_z, w, l, h, \theta)$, where $(p_x, p_y, p_z)$ denotes its center location, $w, l, h$ and $\theta$ denote its width, length, height and yaw angle, respectively. Suppose that the object has a unit-time velocity $\vec{v}=(v_x, v_y)$. The corresponding points-of-interest $(x^t,y^t,z^t)$ sampled from frame $t$ will satisfy the following condition: 
	\begin{equation}
		(x^t - p_x + v_x\cdot \Delta t)^2 + (y^t - p_y + v_y\cdot \Delta t)^2 < (\frac{d^t}{2})^2,
		\label{eq:pool}
	\end{equation}
	where $\Delta t=T-t$ is the time offset of frame $t$ and $d^t = \sqrt{(w^2+l^2)} \cdot \gamma^{\Delta t+1}$ is the diameter of cylindrical region.
	
	In our preliminary study, we compare the overall recall rates of foreground points between our method and the trajectory-based method \cite{mppnet}. As can be seen in Tab.~\ref{tbl:pp}, our method can achieve very close result to the the proposal trajectory method on 4-frame sequences. For 8-frame sequences, since it is difficult to estimate the positions of fast-moving objects on distant frames, we use $\gamma$=1.1 to recall more points and obtain comparable result to the proposal trajectory method. The recall rates only drop slightly even when 16-frame sequences are used. Interestingly, we find that setting $\gamma$=1.1 is not only beneficial for detecting fast-moving objects but also beneficial for improving the performance on slow and stationary objects. We believe this is because the points sampled from regions of different sizes contain multi-level contextual information about the objects, which will be further discussed in Sec.~\ref{sec:ablation}.
	
	\begin{figure*}[t]
		\centering
		\includegraphics[width=0.99\textwidth]{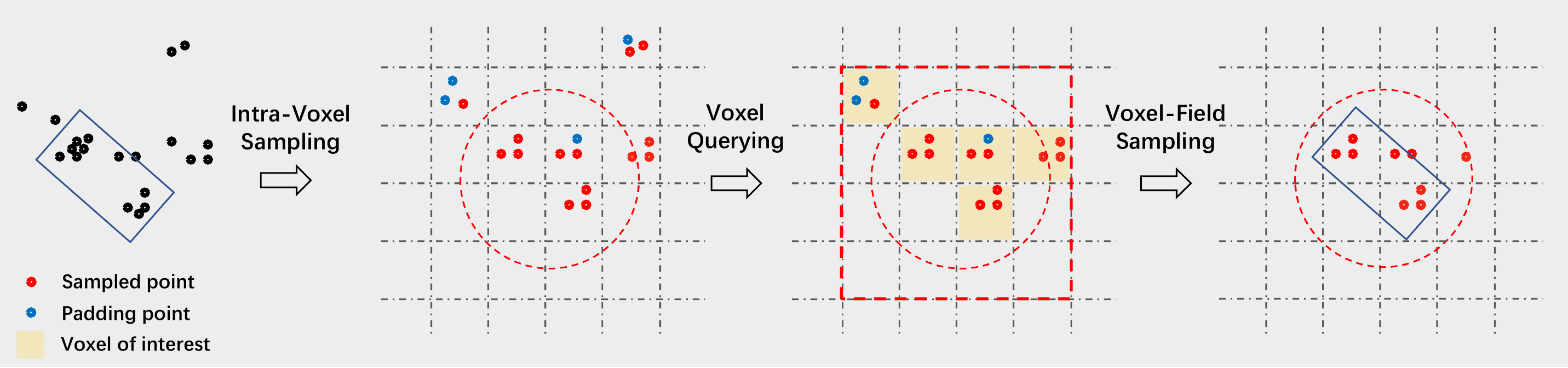}
		\caption{Illustration of our optimized point cloud pooling method. We first perform intra-voxel sampling to keep a fixed number of points in each voxel. Then we query $n\times n$ voxels fields for each proposal and uniformly draw points from the non-empty voxels within.} 
		\label{fig:pool}
	\end{figure*}
	
	\subsection{Region-based Network}
	\textbf{Proposal feature encoding.} After sampling $K$ points in each proposal, we adopt two encoding schemes to generate proposal features.	
	We first follow \cite{ct3d} to calculate the relative offsets between each point-of-interest $l^t_i \in I^t$ and the nine key-points (eight corner points plus one center point) of the proposal box $\{b^t_j \in P^t:j=0,...,8\}$. The resulted offsets are then converted to spherical coordinates and transformed, via a Multi-layer Perceptron (MLP), to a geometric embedding $g^t \in {\rm \mathbb{R}}^{K\times D}$ that encodes the spatial correlation between the sampled point and the proposal box. The encoding scheme can be formulated as:
	\begin{equation}
		g_i^t =\text{MLP}(\mathcal S ( \{l_i^t-b^t_{j}\}_{j=0}^8) ), \text{for } i=1,...,K,
		\label{eq:pbe}
	\end{equation}
    where $\mathcal S:(x, y, z)\rightarrow (r, \theta, \phi)$ denotes the spherical transform where $r=\sqrt{x^2+y^2+z^2}$, $\theta=\arcsin(z/r)$ and $\phi=\arctan(y/x)$ respectively. The second scheme produces a motion embedding $m^t \in {\rm \mathbb{R}}^{K\times D}$, which encodes the displacements of the points-of-interest at each frame $I^t$ relative to the key-points of the proposal boxes $b^0$ in the first frame. The time offsets $\Delta t$ are also concatenated, resulting in the motion embedding at frame $t$ as:
	\begin{equation}
		m_i^t = \text{MLP}(\text{Concat}( \{l_i^t-b^0_{j}\}_{j=0}^8, \Delta t)), \text{for } i=1,...,K.
	\end{equation}
	The proposal feature $f^t \in {\rm \mathbb{R}}^{K\times D}$ can be formulated as the summation of geometric and motion embeddings:
	\begin{equation}
		f^t = g^t + m^t. 
	\end{equation}
	
	\textbf{Bidirectional feature aggregation.} MSF employs a region-based network to explore \textit{spatial-temporal} dependencies among proposal features and fuse them into global representations for final bounding-box refinement.  As shown in Fig.~\ref{fig:model}, the region-based network is composed of three learning blocks.  Each block consists of a traditional Multi-Head Self-Attention (MHSA) layer, followed by a Feed-Forward Network (FFN) with residual connection, and our proposed Bidirectional Feature Aggregation (BiFA) module.  The MHSA layer aims to encode rich \textit{spatial} relationships and point dependencies in the proposal for refining point features, while the proposed BiFA module aims to encode \textit{temporal} information by facilitating information exchange between proposals across frames.
	
	Specifically, BiFA involves a forward path and a backward path, representing two ways of information flow along the sequence. Since proposal features have an unordered representation, we leverage the \textit{Max-pool\&Repeat} \cite{voxelnet} tensor manipulation to obtain a summarized contextual features. In the forward path, aside from the first frame in the sequence, which is concatenated with its own contextual features, each of the other frames is concatenated with the contextual features of its preceding frame along channel dimension. A point-wise convolutional layer is thereby employed to refine the concatenated feature and halve their channels. Given proposal features $f^{t}$ and $f^{t-1}$ from the current frame and its preceding frame, the forward output of frame $t$, denote by $h^t_F$, can be obtained by:
	\begin{align}
		h^{t}_{F} &= \text{Conv} ( \text{Concat}(f^{t},  \text{Repeat} \circ \text{Max-pool}(f^{t-1})) ).
	\end{align}
	
	Unfortunately, introducing the forward path only will lead to information imbalance among different frames, \ie, the current frame receives information from other frames, whereas the last preceding frame receives no information from the sequence. To overcome this limitation, we augment the backward path, where the features of frame $t$ are aggregated with the contextual features of frame $t+1$, resulting in the backward output $h^{t}_B$ as follows:
	\begin{align}
		h^{t}_{B} &= \text{Conv} ( \text{Concat}(h^{t}_{F}, \text{Repeat} \circ \text{Max-pool}(h^{t+1}_{F})) ).
	\end{align}
	In this way, each frame can simultaneously exchange information with its two adjacent frames from both directions, and the information can be propagated to more distant frames after three learning blocks. The resulted proposal features can capture long-term dependencies over the point cloud sequence. It is worthy noting that, the parameters of the convolutional layer in the same path can be shared and all the tensor operations can be performed in parallel. This makes our BiFA module lightweight and much more efficient than other multi-frame fusion modules. 
	
	{\bf Outputs and loss function.} Finally, we aggregate the point-wise features of each proposal into a global representation through a query-based transformer decoder layer, where a learnable query feature $q\in {\rm \mathbb{R}}^D$ attends to each point vector of proposal features $h^t\in {\rm \mathbb{R}}^{K\times D}$ through a cross-attention module. The decoder layer can be depicted as follows:
	\begin{align}
		\hat e^t &= \text{Attention}(q, h^t, h^t) + h^t,\\
		e^t &= \text{FFN}(\hat e^t) + \hat e^t.
	\end{align}
	
	The decoded outputs across frames $\{e^t\}_{t=0}^T$ are concatenated and passed to a group of detection heads for bounding-box refinement. The overall loss function $\mathcal{L}_{\rm total}$ is the summation of the confidence prediction loss $\mathcal L_{\rm conf}$ and the box regression loss $\mathcal L_{\rm reg}$ as follows:
	\begin{equation}
		\mathcal {L}_{\rm total}=\mathcal {L}_{\mathrm {conf}}+ \alpha \mathcal {L}_{\mathrm {reg}}, \label {eq:loss}
	\end{equation}
	where $\alpha$ is a balancing hyper-parameter. We adopt the same binary cross entropy loss and box regression loss employed in CT3D \cite{ct3d} as our $\mathcal L_{\rm conf}$ and $\mathcal L_{\rm reg}$.

	\subsection{Efficient Point Cloud Pooling}
	In this subsection, we optimize the point cloud pooling method to sample a fixed number of points more efficiently from the cylindrical region of each proposal. As shown in Fig.~\ref{fig:pool}, we perform proposal-based point sampling in two steps, \ie, the intra-voxel sampling and the voxel-field sampling. In the first step, the input space is discretized into a voxel grid with voxel size $v$, and each point corresponds to a voxel coordinates $(\floor{\frac{x}{v}}, \floor{\frac{y}{v}})$. Here the voxel partition in z-axis is omitted considering that the cylindrical regions have unlimited height. Then, up to $k$ points are kept in each voxel and padding points are inserted if the voxel has less than $k$ points. Due to the high memory consumption required to store all voxels in a dense grid, we follow \cite{votr} to store only non-empty voxels in a contiguous memory and use a hash table to store the mappings between the  coordinates of non-empty voxels and their indices in memory space.

	\begin{table}[t]
		
		\caption{The latency of point cloud pooling on 1-frame, 4-frames and 8-frames sequences.}
		\vspace{-2mm}
		\centering
		\begin{adjustbox}{width=0.9\columnwidth}
			\begin{tabular}{c|c|c|c}
				\hline

				{}  & \textit{N}=168k  & \textit{N}=674k   & \textit{N}=1382k \\ \hline
				Cylindrical Pooling & 8.2ms & 25.2ms &40.1ms\\
				Our Optimized  &2.3ms &3.4ms&5.0ms \\
				\hline
				
			\end{tabular}
		\end{adjustbox}
		\label{tbl:pool}
		\vspace{-3mm}
	\end{table}
	
	In the second step, we first query $n$-by-$n$ voxel-field for each proposal and calculate the coordinates of the voxels within. Next, we convert the coordinates of voxel into hashed keys and look up the table for querying the sampled points generated from the first hierarchy. The hash-table will return ``-1'' if no keys are found, which means it is an empty voxel. The queried points are then drawn from these voxels and stored in an output buffer if the following conditions are met: 1) it is a valid point and 2) it falls into the cylindrical region, \ie satisfying Eq. \ref{eq:pool}.

	\textbf{Complexity and efficiency}
	We perform an in-depth analysis of our optimized pooling method and the previous  cylindrical pooling methods \cite{mppnet, ct3d, std} in continuous space. Given $N$ points, $M$ proposals and the requirement of sampling $K$ points in each proposal, the original method costs $\mathcal O(NM) + \mathcal O(MK)$ complexity due to the calculation of pair-wise distances between points and proposals. In contrast, our optimized version costs $\mathcal O(N) + \mathcal O(M) + \mathcal O(MK)$ complexity, where the first term is the cost of intra-voxel sampling, the second term is the cost of querying voxel-field for each proposal and the third term is the cost of drawing points from the queried voxels. We evaluate the latency of point cloud pooling on sequences with different lengths. As shown in Tab.~\ref{tbl:pool}, our optimized pooling method can achieve an 8$\times$ speedup over the original pooling method.

	\begin{table*}[t]
		\caption{Performance comparison on the validation set of Waymo Open Dataset.}
		\centering
		\begin{adjustbox}{width=1\textwidth}
			\begin{tabular}{c|c|cc|cc|cc|cc}
				\toprule[2pt]
				
				\multirow{2}{*}{Method} &
				\multirow{2}{*}{Frames} &
				\multicolumn{2}{c|}{ALL (3D mAPH)  } &
				\multicolumn{2}{c|}{Vehicle (AP/APH)  } &
				\multicolumn{2}{c|}{Pedestrian (AP/APH)   } &
				\multicolumn{2}{c}{Cyclist (AP/APH)  } \\
				
				&  &L1 & L2& L1& L2& L1& L2& L1& L2\\ 
				\toprule[2pt]
				
				SECOND \cite{second} & 1 &63.05& 57.23 &72.27/71.69& 63.85/63.33& 68.70/58.18& 60.72/51.31& 60.62/59.28& 58.34/57.05 \\
				PointPillar \cite{pointpillars}& 1& 63.33& 57.53& 71.60/71.00& 63.10/62.50& 70.60/56.70& 62.90/50.20& 64.40/62.30& 61.90/59.90\\
				IA-SSD \cite{ia-ssd} & 1& 64.48&58.08&70.53/69.67& 61.55/60.80 &69.38/58.47& 60.30/50.73& 67.67/65.30 &64.98/62.71\\
				LiDAR R-CNN \cite{lidar-rcnn}& 1& 66.20& 60.10& 73.50/73.00& 64.70/64.20& 71.20/58.70& 63.10/51.70& 68.60/66.90& 66.10/64.40\\
				RSN \cite{rsn}& 1& -& -& 75.10/74.60& 66.00/65.50& 77.80/72.70& 68.30/63.70& -& -\\

				PV-RCNN \cite{pvrcnn}& 1& 69.63& 63.33& 77.51/76.89& 68.98/68.41& 75.01/65.65& 66.04/57.61& 67.81/66.35& 65.39/63.98\\
				Part-A2 \cite{part-aware}& 1 &70.25& 63.84& 77.05/76.51& 68.47/67.97& 75.24/66.87& 66.18/58.62& 68.60/67.36& 66.13/64.93\\
				Centerpoint \cite{centerpoint} &1& -& 65.50& -& -/66.20& -& -/62.60& -& -/67.60\\
				
				VoTR \cite{votr} & 1 & -&-&74.95/74.25&65.91/65.29&-&-&-&-\\
				VoxSeT \cite{voxset} & 1 & 72.24 & 66.22 &74.50/74.03&65.99/65.56&80.03/72.42&72.45/65.39&71.56/70.29&68.95/67.73\\
				SST-1f\cite{sst}&1& -& -&76.22/75.79& 68.04/67.64&81.39/74.05& 72.82/65.93&-&-\\
				SWFormer-1f\cite{swformer} &1& -& -&77.8/77.3& 69.2/68.8&80.9/72.7 &72.5/64.9 &-&- \\
				PillarNet \cite{pillarnet}& 1& 74.60&68.43&79.09/78.59& 70.92/70.46& 80.59/74.01& 72.28/66.17& 72.29/71.21& 69.72/68.67\\
				PV-RCNN++ \cite{pvrcnn++}&1& 75.21& 68.61& 79.10/78.63& 70.34/69.91& 80.62/74.62& 71.86/66.30& 73.49/72.38& 70.70/69.62\\

				\toprule[2pt]
				
				3D-MAN \cite{3dman} &16& -& -& 74.53/74.03& 67.61/67.14& 71.7/67.7& 62.6/59.0&-&-\\
				SST-3f\cite{sst}&3& -& -&78.66/78.21& 69.98/69.57&83.81/80.14& 75.94/72.37&-&-\\
				SWFormer-3f\cite{swformer}&3& -& -&79.4/78.9& 71.1/70.6&82.9/79.0& 74.8/71.1&-&-\\
				CenterFormer \cite{centerformer} & 4 & 77.0 & 73.2 &78.1/77.6& 73.4/72.9& 81.7/78.6& 77.2/74.2& 75.6/74.8& 73.4/72.6\\  
				CenterFormer \cite{centerformer} & 8 & 77.3 & 73.7 &78.8/78.3& 74.3/73.8& 82.1/79.3& 77.8/75.0& 75.2/74.4& 73.2/72.3 \\ 
				MPPNet \cite{mppnet} &4& 79.83& 74.22& 81.54/81.06& 74.07/73.61& 84.56/81.94& 77.20/74.67& 77.15/76.50& 75.01/74.38\\
				MPPNet \cite{mppnet} &16& 80.40& 74.85& 82.74/\textbf{82.28}& 75.41/74.96& 84.69/82.25& 77.43/75.06& 77.28/76.66& 75.13/74.52\\
				
				\toprule[2pt]
				MSF (ours) & 4 & 80.20 & 74.62 & 81.36/80.87& 73.81/73.35& 85.05/82.10& 77.92/75.11& 78.40/77.61& 76.17/75.40\\
				MSF (ours) & 8 & \textbf{ 80.65} & \textbf{ 75.46} & \textbf{ 82.83}/82.01& \textbf{75.76/75.31}& \textbf{ 85.24/82.21}& \textbf{ 78.32/75.61}& \textbf{ 78.52/77.74}& \textbf{ 76.32/75.47}\\
				\toprule[2pt]
			\end{tabular}
		\end{adjustbox}
		\label{tbl:val}
		\vspace{-3mm}
	\end{table*}

	\section{Experiment}
	\subsection{Dataset and Implementation Details}
	\textbf{Dataset.}
	We evaluate our MSF on Waymo Open Dataset (WOD), which is a large-scale multi-modality dataset for autonomous driving. WOD contains 1,150 sequences, which are divided into 798 training,  202 validation, and 150 testing sequences, respectively. Each sequence is 20s long, captured by a 64-line LiDAR sensor at 10Hz frequency.  The evaluation metrics used in WOD are mean average precision (mAP) and mAP weighted by heading accuracy (mAPH). Three object classes, "Vehicle", "Pedestrian" and "Cyclist", are evaluated. Each object class is further categorized into two levels of difficulties, LEVEL1 and LEVEL2. The former refers to objects with more than 5 points and the latter refers to objects with less than 5 but at least 1 point.
	
	\textbf{Implementation.}
	We employ the traditional CenterPoint \cite{voxelnet, second, centerpoint} as our region proposal network (RPN). To include motion information, we concatenate four adjacent frames at input and add one additional head for predicting the velocity of objects. In our experiments, we first train RPN using the official settings of OpenPCDet\footnote{https://github.com/open-mmlab/OpenPCDet/}, and use it to generate proposals of WOD. Based on these proposals, we then train our region-based network for 6 epochs by using the ADAM optimizer with an initial learning rate of 0.003 and a batch size of 16. The learning rate is decayed with One-Cycle policy and the momentum is set between [85\%, 95\%]. During the training, we sample region proposals with IoU$>$0.5 and conduct proposal augmentation following PointRCNN \cite{pointrcnn}. A number of 128 raw LiDAR points are randomly sampled for each proposal. The voxel size $v$ and the points-per-voxel $k$ used in voxel-based sampling are set to 0.4 and 32, respectively. The feature dimension of each point in the learning block is set to 256. At the training stage, we use the intermediate supervision by adding loss to the output of each learning block and sum all the intermediate losses to train the model. At the test stage, we only use the bounding boxes and the confidence scores predicted from the last learning block.
	
	\begin{table*}[t]
		\caption{Performance comparison on the test set of Waymo Open Dataset.}
		\centering
		\begin{adjustbox}{width=1\textwidth}
			\begin{tabular}{c|cc|cc|cc|cc}
				\toprule[2pt]
				
				\multirow{2}{*}{Method} &
				
				\multicolumn{2}{c|}{ALL (3D mAPH)  } &
				\multicolumn{2}{c|}{Vehicle (AP/APH)  } &
				\multicolumn{2}{c|}{Pedestrian (AP/APH)   } &
				\multicolumn{2}{c}{Cyclist (AP/APH)  } \\
				
				&L1 & L2& L1& L2& L1& L2& L1& L2\\ 
				\toprule[2pt]
				PointPillar\cite{pointpillars}&-&-&68.10& 60.10& 68.00/55.50& 61.40/50.10&-&-\\
				StarNet\cite{starnet}&-&-&61.00& 54.50& 67.80/59.90& 61.10/54.00 &-&-\\
				M3DETR\cite{m3detr} & 67.1 &61.9 &77.7/77.1& 70.5/70.0& 68.2/58.5& 60.6/52.0& 67.3/65.7& 65.3/63.8\\ 
				3D-MAN \cite{3dman} & -& -& 78.28& 69.98& 69.97/65.98& 63.98/60.26& -& -\\
				PV-RCNN++ \cite{pvrcnn++} & 75.7& 70.2&81.6/81.2& 73.9/73.5& 80.4/75.0& 74.1/69.0& 71.9/70.8& 69.3/68.2 \\
				CenterPoint \cite{centerpoint}& 77.2& 71.9& 81.1/80.6 &73.4/73.0& 80.5/77.3& 74.6/71.5& 74.6/73.7& 72.2/71.3 \\
				RSN \cite{rsn} & -& -& 80.30& 71.60& 78.90/75.60 &70.70/67.80 &-&-\\
				SST-3f \cite{sst} &78.3 &72.8 &81.0/80.6 &73.1/72.7& 83.3/79.7 &76.9/73.5 &75.7/74.6 &73.2/72.2 \\
				MPPNet \cite{mppnet}& 80.59& 75.67 & 84.27/83.88 &77.29/76.91& 84.12/81.52& 78.44/75.93& 77.11/76.36& 74.91/74.18\\
				CenterFormer \cite{centerformer} & 80.91& 76.29& 85.36/84.94 &78.68/78.28 & 85.22/	82.48&80.09/77.42&76.21/75.32&74.04/73.17 \\
				
				\toprule[2pt]
				MSF (ours) &\textbf{81.74}&\textbf{76.96}& \textbf{86.07/85.67}&\textbf{79.20/78.82} & \textbf{85.99/83.10}&\textbf{80.61/77.82}&\textbf{77.29/76.44}&\textbf{75.09/74.25}\\
				
				\toprule[2pt]
			\end{tabular}
		\end{adjustbox}
		\label{tbl:test}
		
	\end{table*}
	
	\begin{table}[t]
		\centering
		\caption{Ablation experiments on the WOD validation set. ``ME'' refers to using Motion Embedding and ``SA'' refers to using Self-Attention modules. ``Q'' means using query-based decoder layer to generate global representation of proposal features and ``M'' means using single max-pooling layer for global feature generation. APH scores on LEVEL1 and LEVEL2 are reported. }
		\begin{adjustbox}{width=1\columnwidth}
			\begin{tabular}{cccc|ccc}
				\hline

				ME & SA & BiFA & Decoder  & L1 & L2   \\ \hline\hline
				\checkmark & \checkmark &  	\checkmark & Q & 80.20 & 74.62  \\
				\ding{56}		   & \checkmark & 	\checkmark 	& Q & 80.03 (\textcolor{blue}{-0.17})  & 74.50 (\textcolor{blue}{-0.12})   \\
				
				\checkmark & 	 \ding{56}				& \checkmark & Q & 76.51 (\textcolor{blue}{-3.69})  & 71.91 (\textcolor{blue}{-2.71})  \\
				\checkmark & \checkmark &  	 \ding{56}				 & Q & 78.25 (\textcolor{blue}{-1.95}) & 73.11 (\textcolor{blue}{-1.51})  \\

				\checkmark & \checkmark & \checkmark & M & 79.54 (\textcolor{blue}{-0.66})   & 74.08 (\textcolor{blue}{-0.54})  \\
				
				\hline
			\end{tabular}
		\end{adjustbox}
		\label{tbl:abl}
		
	\end{table}

	\begin{table}[t]
		\centering
		\caption{Bidirectional feature aggregation vs. unidirectional feature aggregation.}
		\begin{adjustbox}{width=0.9\columnwidth}
			\begin{tabular}{cc|cc}
				\hline

				Forward & Backward  & L1 & L2   \\ \hline\hline
				\checkmark & \checkmark &   80.20 & 74.62  \\
				\checkmark	& \ding{56} & 	79.64 (\textcolor{blue}{-0.56}) & 74.35 (\textcolor{blue}{-0.27}) \\ 
				\ding{56}	& \checkmark & 	78.97 (\textcolor{blue}{-1.23}) & 73.77 (\textcolor{blue}{-0.85}) \\ 
				\hline
			\end{tabular}
		\end{adjustbox}
		\label{tbl:bifa}
		
	\end{table}

	\begin{table}[t]
		\centering
		\caption{The performance by integrating the proxy points and MLP mixer from MPPNet. ``SA'' and ``Mixer'' denote the self-attention and MLP Mixer \cite{mppnet}, respectively. The APH scores of LEVEL2 on three object classes are reported.}
		\begin{adjustbox}{width=1\columnwidth}
			\begin{tabular}{c|ccc}
				\hline
				
				Config  & Vehicle.  & Pedestrian. & Cyclist  \\ \hline\hline
				Raw + SA & 73.35 & 75.11 & 75.40  \\
				Proxy + SA & 73.12 (\textcolor{blue}{-0.23}) & 74.20 (\textcolor{blue}{-0.91}) & 74.32 (\textcolor{blue}{-1.08}) \\
				Proxy + Mixer & 73.45(\textcolor{blue}{+0.10}) & 74.13 (\textcolor{blue}{-0.98})& 74.39 (\textcolor{blue}{-1.01}) \\ 
				
				\hline
			\end{tabular}
		\end{adjustbox}
		\label{tbl:mppnet}
		
	\end{table}
	
	\subsection{Results on WOD}
	\textbf{Waymo Validation Set.} Tab.~\ref{tbl:val} compares the performance of MSF and the current state-of-the-art methods. As can be seen, multi-frame detectors generally outperform single-frame methods. The best two multi-frame methods by far are MPPNet \cite{mppnet} based on proposal trajectory and CenterFormer \cite{centerformer} based on transformer model.	Using the same number of frames, our MSF method significantly outperforms CenterFormer by 3.4\% APH and 1.2\% APH on LEVEL1 and LEVEL2, respectively. This demonstrates that fusing multiple frame feature at the region level is more effective than the fusion with convolutional features. Compared with MPPNet \cite{mppnet}, which is also based on region-level fusion, our method outperforms it on almost all cases, except for the APH of Vehicle LEVEL1. It should be noted that our MSF method only use 8 frames, while MPPNet uses 16 frames. MSF uses different pooling sizes in different frames, therefore generating multi-level contextual features for each object. Specifically, MSF models with 4- and 8-frames achieve the mAPH of 74.62\% and 75.13\%, respectively, recording new state-of-the-arts. In addition, both CenterFormer and MPPNet extract features on each frame of the sequence with the need of a memory bank to store the intermediate results of preceding frames, while our method performs proposal generation only on the current frame, which is more practical to online inference.
	
	\textbf{Waymo Test Set.} In Tab.~\ref{tbl:test}, we show the results of our 8-frame model by submitting the prediction result to the online server for evaluation. Our method outperforms all the previously published methods. In particular, the improvements on Vehicle and Pedestrian classes are significant, outperforming the best competitor, \ie, CenterFormer, by 1.91\% APH and 1.89\% APH on LEVEL2, respectively.

	\subsection{Ablation Studies}
	\label{sec:ablation}
	In this section, we conduct in-depth analysis on MSF by evaluating the effectiveness of each individual component of it. We report the APH metric of our 4-frame model on the WOD validation set. The ablation study results are shown in Tab.~\ref{tbl:abl} to Tab.~\ref{tbl:gamma}, respectively. 
	
	\textbf{Motion embedding.}
	As can be seen from the \nth{1} and \nth{2} rows of Tab.~\ref{tbl:abl}, without motion encoding, the performance of our proposed MSF will drop by 0.17\% on LEVEL1 and 0.12\% in LEVEL2. This indicates that the motion information is beneficial to infer the object's geometry information in point cloud sequences.
	
	\textbf{Self-attention.} As can be seen from the \nth{1} and \nth{3} rows of Tab.~\ref{tbl:abl}, without using self-attention for spatial interactions, MSF will suffer from significant performance drop, 3.69\% on LEVEL1 and 2.71\% on LEVEL2. This indicates that the intra-frame interaction is vital for learning internal geometry information of the proposals. 
	
	\textbf{Bidirectional feature aggregation.} From the \nth{1} and \nth{4} rows of Tab.~\ref{tbl:abl}, we can find that enforcing temporal interactions among hidden features of proposals by using the proposed BiFA module can bring an improvement of 1.95\% on LEVEL1 and 1.51\% on LEVEL2, which demonstrates the benefits of exploring long-term dependencies in the point cloud sequence.  We also conduct experiments by performing unidirectional feature aggregation with either forward or backward path. The results are shown in Tab.~\ref{tbl:bifa}, from which we see that the unidirectional model obtains lower APH scores than that of the bidirectional model.

\textbf{Query-based decoder layer.} As shown in the last row of Tab.~\ref{tbl:abl}, by replacing the final query-based decoder layer with a single max-pooling layer, the performance will drop slightly by 0.66\% and 0.54\%. This is because the final proposal representation after decoder can be regarded as a weighted sum of all point features, and the decoding weights of different points can intrinsically introduce more dynamics for feature representation.
	
	\textbf{Spatial modeling with proxy points and MLP mixer}. 
	We perform an analysis by integrating the proxy points and MLP mixer developed in our best competitor MPPNet \cite{mppnet} into MSP. We first follow MPPNet to generate proxy point of each proposal and apply a set abstraction \cite{pointnet++} to aggregate the point-wise features to every proxy point. As shown in the \nth{1} and \nth{2} rows of Tab.~\ref{tbl:mppnet}, using features of proxy points will degrade the performance on Vehicle and more significantly on Pedestrian and Cyclist classes. After replacing the self-attention module with the MLP mixer, the performance degradation on Pedestrian and Cyclist classes still exists, as shown in the \nth{3} row of Tab.~\ref{tbl:mppnet}. This phenomenon is contradictory to the conclusion in MPPNet that using proxy points to formulate proposal features is more effective than using raw points. We believe this is because MPPNet applies per-frame detection, and thus the proposals across frames may have different sizes, while proxy points can provide consistent representations for different frames. In contrast, our method uses propagated proposals with the same size over the sequence, and hence our proposal features are naturally on the same scale. For small objects such as pedestrian and cyclist, the features from raw points can provide more fine-grained details than proxy points. 
	
	\textbf{Effects of $\gamma$}. We evaluate how $\gamma$ affects the performance on the objects with different speeds. As shown in Tab.~\ref{tbl:gamma}, gradually expanding the proposal region with $\gamma$=1.1 will bring substantial improvements on the moving objects. This demonstrates that our model is able to capture the information of fast moving objects, even in the distant frames. Interestingly, we also see improvements on slow and stationary objects. This is because the different pooling sizes could provide multi-level contextual information, which is beneficial for detecting the objects. As shown in the \nth{3} row of Tab.~\ref{tbl:gamma}, no further improvements can be made by increasing the value of $\gamma$ to 1.2.

	\begin{table}[t]
		\centering
		\caption{The APH (L2) scores on objects with different velocities using the 8-frame model. The objects are categorized into stationary ($<$0.2m/s), slow (0.2$\sim$1m/s), medium (1$\sim$6m/s) and fast ($>$6m/s) groups. }
		\begin{adjustbox}{width=0.9\columnwidth}
			\begin{tabular}{c|cccc}
				\hline
				
				& Stationary  & Slow & Medium & Fast  \\ \hline\hline		
				$\gamma$=1.0 & 70.32 & 70.30  & 75.60  & 80.32 \\
				$\gamma$=1.1 & 70.43 &  71.17 & 77.56 & 82.48 \\ 
				$\gamma$=1.2 & 70.35 &   71.23 & 77.44 &  82.31 \\ 
				\hline
			\end{tabular}
		\end{adjustbox}
		\label{tbl:gamma}
		\vspace{-5mm}
	\end{table}

	\subsection{Runtime Analysis}
	
\begin{figure}[t]
		\includegraphics[width=1\columnwidth]{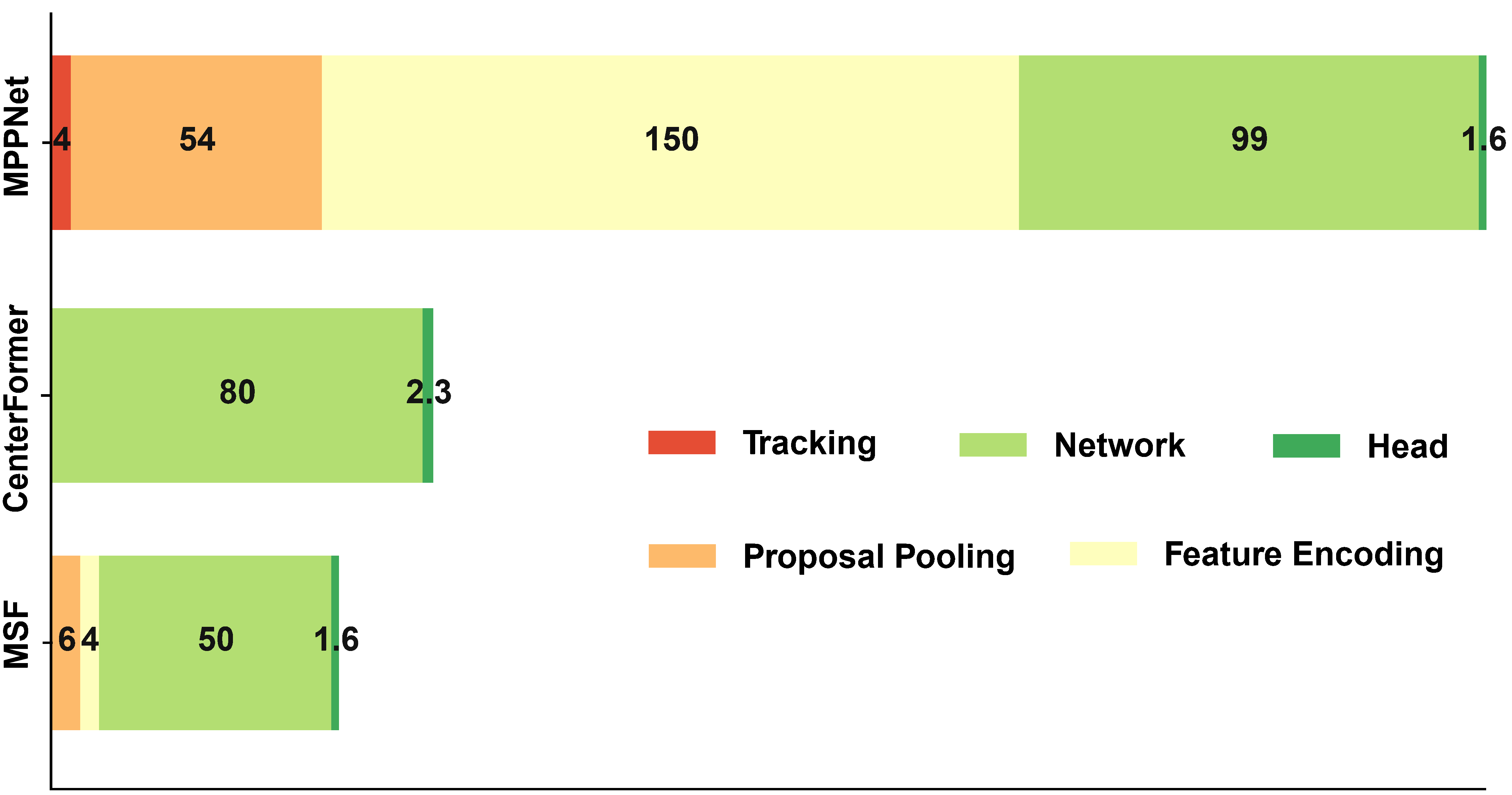}
		
		\caption{ Comparison of the runtime of different methods. 
		}
		\label{fig:runtime}
		
	\end{figure}

Fig.~\ref{fig:runtime} illustrates the latency decomposition of different methods.  Here the latency is computed as the average runtime over 100 samples that are randomly drawn from WOD validation set, which is measured by a single Nvidia GeForce RTX A6000 GPU.
Since all the methods employ the same backbone network, we exclude the runtime of the backbone from the latency for better comparison. For MPPNet and CenterFormer, we assume that the features from the preceding frames are already available in the memory bank. As can be seen, MPPNet costs much time in the feature encoding stage because it performs set-abstraction \cite{pointnet++} to group raw points on the proxy points, which requires manipulating point clouds in continuous space. In comparison, our method takes only 4 milliseconds to encode raw points as proposal features. The overall latency of MPPNet, CenterFormer and MSF are 99, 80 and 50 milliseconds, respectively. Tab.~\ref{tbl:runtime} further decomposes the network latency regarding the spatial and temporal modules in MPPNet and MSF. As can be seen, the self-attention module and BiFA module are much more efficient than MLP-Mixer and cross-attention module, respectively. Thanks to our optimized point cloud pooling, MSF achieves higher efficiency even compared with CenterFormer, which performs feature fusion directly on the convolutional features.
	
	\begin{table}[t]
		\centering
		\caption{The runtime decomposition of MPPNet and MSF. }
			\vspace{-2mm}
		\begin{adjustbox}{width=1\columnwidth}
			\begin{tabular}{cc|cc}
				\hline
				\multicolumn{2}{c|}{MPPNet}  & \multicolumn{2}{c}{MSF} \\
				\cline{1-4}
				MLP-Mixer & Cross-Attention & Self-Attention  & BiFA  \\ 
				75 ms  & 24 ms   & 36 ms & 12 ms\\ 
				
				\hline
			\end{tabular}
		\end{adjustbox}
		\label{tbl:runtime}
		\vspace{-3mm}
	\end{table}

	\section{Conclusion}
	We presented a novel motion-guided sequential fusion method, namely MSF, for 3D object detection from point cloud sequence. Unlike previous multi-frame detectors that performed feature extraction on each frame of the sequence, MSF adopted a proposal propagation strategy to mine points-of-interest based on the proposals generated on the current frame, therefore reducing the redundant computations and relieving reliance on the preceding results. A bidirectional feature aggregation module was proposed to enable cross-frame interaction between proposal features, facilitating MSF to capture long-term dependencies over the sequence. Besides, we optimized the point cloud pooling process, allowing MSF to process large-scale point cloud sequences in milliseconds. The proposed MSF achieved state-of-the-art performance on the Waymo Open Datasets and it was more efficient than other multi-frame detectors. In future research, we plan to extend MSF to generate detection priors on the future frames to further reduce the overall computation of multi-frame detection.

	{\small
		\bibliographystyle{ieee_fullname}
		\bibliography{egbib}
	}
	
\end{document}